# Mitigating Clinician Information Overload: Generative AI for Integrated EHR and RPM Data Analysis

Ankit Shetgaonkar
ankiit@google.com

Lakshit Arora
lakshit@google.com

Shashank Kapoor
shashankkapoor@google.com

Dipen Pradhan
dipenp@google.com

Sanjay Surendranath Girija
sanjaysg@google.com

Aman Raj
amanraj@google.com

Google

*Abstract* — Generative Artificial Intelligence (GenAI), particularly Large Language Models (LLMs), offer powerful capabilities for interpreting the complex data landscape in healthcare. In this paper, we present a comprehensive overview of the capabilities, requirements and applications of GenAI for deriving clinical insights and improving clinical efficiency. We first provide some background on the forms and sources of patient data, namely real-time Remote Patient Monitoring (RPM) streams and traditional Electronic Health Records (EHRs). The sheer volume and heterogeneity of this combined data present significant challenges to clinicians and contribute to information overload.

In addition, we explore the potential of LLM-powered applications for improving clinical efficiency. These applications can enhance navigation of longitudinal patient data and provide actionable clinical decision support through natural language dialogue. We discuss the opportunities this presents for streamlining clinician workflows and personalizing care, alongside critical challenges such as data integration complexity, ensuring data quality and RPM data reliability, maintaining patient privacy, validating AI outputs for clinical safety, mitigating bias, and ensuring clinical acceptance. We believe this work represents the first summarization of GenAI techniques for managing clinician data overload due to combined RPM / EHR data complexities.

## I. Introduction

Modern healthcare is characterized by an unprecedented confluence of patient data from multiple sources. Clinicians are often required to synthesize combined information from traditional Electronic Health Records (EHRs) with rapidly growing Remote Patient Monitoring (RPM) data. EHR data contains repositories of largely episodic data captured in clinical settings [28]. RPM data consists of continuous and real-world data generated by Remote Patient Monitoring programs utilizing wearable biosensors and other connected health platforms [5][15]. There is a rapid growth in data, particularly RPM data, covering everything from vital signs and activity levels to glucose trends obtained outside of clinical settings. This presents both significant opportunities and major challenges. Current clinical workflows rely on manual reviews, using dashboards that typically present a fragmented view of the data [1][20]. This increases the risk of information overload, cognitive burden and burnout for clinicians, which can affect the quality and timeliness of care [36].

The advent of GenAI [11], particularly LLMs, offers a transformative opportunity to navigate this data overload and complexity [9]. Advanced models like GPT-4, Med-PaLM 2, and Gemini demonstrate remarkable capabilities in understanding natural language [4], processing diverse information formats including text, images and representations of time-series data common in RPM [33], performing complex reasoning [15], and generating human-like responses [27]. The proficiency of LLMs in human dialogue enables the development of intuitive conversational interfaces [34] for clinicians to potentially interact with complex, integrated patient datasets using natural language queries. Some possible applications are: requesting summaries of longitudinal RPM trends contextualized by EHR events, exploring correlations, or asking specific questions in a unified conversational session without needing specialized analytical tools or manually collating information from disparate sources.

In this paper, we provide a comprehensive overview of leveraging GenAI, especially LLMs, to manage the complexity arising from integrated EHR and RPM data. We examine how the volume and heterogeneity of this combined data contribute to clinician information overload. We explore the potential of LLM-powered applications for clinicians to navigate this data through natural language and improve clinical efficiency. We also detail the critical challenges that must be addressed, such as data integration, quality assurance, privacy, AI validation, and clinical acceptance. With this review, we aim to guide future innovation in developing clinically relevant decision support tools that effectively leverage this complex data landscape. The organization of this paper is as follows. In Section II, we provide an overview of the patient data landscape, detailing the distinct characteristics of patient data originating from RPM systems and traditional EHRs and its modalities. Section III explores the core capabilities of GenAI, particularly Large Language Models (LLMs), the critical prerequisites for effective and safe application of GenAI in healthcare, and key application areas where GenAI can provide clinical value. Finally, Section IV presents a discussion of the potential implications of a GenAI-based approach, implementation challenges, and future directions for leveraging GenAI to manage complex patient data and support clinical workflows.

## II. Patient Data

Patient data includes a broad spectrum of information about the current health status and medical history of a patient. This data acts as the foundation upon which healthcare providers form clinical decisions and treatment plans. Patient data has multiple inherent modalities and is collected from a large variety of disparate sources. In this section we discuss two main types of patient data: Patient-Generated Health Data [1], typically generated through RPM [1] tools and EHRs containing data captured by clinicians and labs [42][43].

**Information overload** is a growing challenge for healthcare providers in the context of patient data. Clinicians are tasked with deciphering a large amount of digitized information, often overwhelming their cognitive capacity and limiting their ability to process all of the available information effectively for patient care. Some of the challenges here include difficulty in finding, processing, and acting upon relevant patient data due to a large volume of potentially clinically irrelevant or duplicative data, compounded with poor organization or poor user interfaces in EHR systems [41]. More recently, data obtained directly from patients in a non-clinical setting, termed as Patient-generated Health Data, is proving to be a supplementary source of patient data to EHR. The high volume and high velocity of this data from wearable sensors can exacerbate this overload.

The added effort of data entry into EHR systems is a task that adds to cognitive overload for clinicians [41], due to workflows that enforce documentation requirements, sometimes due to regulatory requirements such as the HITECH regulations [57]. Kroth et al (2019) [41] describe this as "note bloat" / "note overload" [56] driven by requirements for long progress notes and additional note taking towards billing, quality improvement measures, avoiding malpractice, compliance, visit history documentation, and physical exam documentation. These tasks are frequently hindered by poor EHR design that detracts from patient care due to misalignment with clinician workflows. [44][45]

### A. Patient-Generated Health Data (PGHD)

Patient-Generated Health Data refers to health-related information that is created, recorded, gathered, or inferred by patients, their families, or caregivers outside of traditional clinical settings [1]. This data can be gathered through a variety of technologies and methods, including digital biosensors, manual log or journal entries.

RPM is a method of healthcare delivery that relies on digital technologies to collect PGHD such as vital signs, weight, blood glucose levels, symptoms, etc. from patients in a remote location, away from traditional clinical settings. PGHD data is electronically transmitted to healthcare providers in a different location for assessment and potential intervention [5]. In the following sections we will discuss the types of RPM data sources.

*1) Wearable Biosensors*

Wearable biosensors include a diverse range of electronic devices which can be worn on, adhered onto or minimally implanted under the skin. These electrophysiological and electrochemical sensors are designed to continuously or frequently capture physiological, behavioral, or biochemical data directly from the individual [5][38].

A few common sensor types used for RPM are: Inertial Measurement Units (IMUs) and pressure sensors [19] to track movement for enabling step counting, activity classification, gait analysis, and fall detection. Optical Photoplethysmography (PPG) sensors and Electrocardiography (ECG) sensors for tracking cardiovascular health. Other physiological measurements include body temperature sensors, Continuous Glucose Monitors (CGMs) and upcoming technologies to monitor biomarkers in breath (Volatile Organic Compounds for metabolic/respiratory status) and sweat (electrolytes, metabolites) [5][36].

Continuous monitoring of these biological parameters through wearable sensor devices offers unprecedented potential for RPM using longitudinal trends in health data, and provides opportunities for personalized interventions, and early disease detection [5][35].

*2) Patient Reported Outcomes (PROs)*

Patient Reported Outcomes include data gathered through surveys, interviews and data capture platforms where patients subjectively report on their health improvements, severity of symptoms, and general health status [21]. Examples include:

- **Logs / Diaries** such as food diaries, pain journals, seizure logs, migraine tracking.
- **Electronic / Mobile Health (eHealth/mHealth) applications** used by patients to self-report various types of health information such as symptoms, medication adherence, lifestyle data (diet, exercise), and mood through websites or mobile apps. Patient Portals provided by healthcare providers enable information to be captured directly from manual entries made by patients or their caregivers.

### B. Electronic Health Records (EHRs)

EHRs are digitized versions of patient charts [28] stored as temporal data in software-based platforms. This data forms longitudinal health records and contains structured (tabular) and unstructured (text) events. Clinicians interact with this information through Graphical User Interfaces (GUI).

EHR systems include data such as:

- **Documentation/Notes** entered by clinicians containing both structured tabular data and unstructured text documentation of a patient's care and treatment
- **Lab Test Results** containing structured data with numerical values, units, reference ranges, flags for

- abnormal results, transmitted electronically through a Laboratory Information System (LIS) to the EHR.
- **Medical Imaging Reports** containing image data and interpretation of the results by a specialist, electronically transmitted from Radiology Information Systems (RIS) and Picture Archiving and Communication Systems (PACS) [17].

*C. Modalities in Patient Data*

As illustrated in the previous sections, patient data originates from various clinical and remote sources, and is present in multiple formats. This inherent heterogeneity and fragmentation present challenges for integration and analysis [10]. Table I provides a consolidated overview of the data modalities in patient data, with examples of data sources and the characteristics of the data as it relates to healthcare applications.

Table I: Health care data modalities

| Modality | Examples |
| --- | --- |
| Text (Unstructured) | Clinical Notes, Radiology Reports, Patient Correspondence |
| Structured Data (Tabular/Coded) | Lab Results, Vital Signs, Medication Lists, Diagnostic Codes, Allergy Lists, Demographics, PROs (Survey responses, diaries) |
| Signal / Waveform (Time-Series) | ECG, PPG, CGM, Blood Pressure, IMU Movement data |
| Image | X-rays, CT scans, MRI, Ultrasound, Pathology Slides |
| Video | Surgical Recordings, Endoscopy, Ultrasound Videos, Telemedicine Sessions |

In the next section, we explore recent research into the use of GenAI applications to alleviate overload while interfacing with patient data.

### III. GENERATIVE AI FOR CLINICAL INSIGHTS FROM INTEGRATED RPM AND EHR DATA: CAPABILITIES, PREREQUISITES AND APPLICATIONS

GenAI represents a significant advancement in the capabilities of AI. GenAI models are capable of generating novel content that mirrors patterns learned from existing data. Bahn and Strobel (2023) [11] showcase different types of GenAI and their abilities in interaction with different modalities of data.

Applications of GenAI in healthcare environments is an area that continues to be the subject of ongoing research and development and has great potential to assist clinicians along with optimizing health care processes [51]. Transformer-based models, which are pre-trained on vast datasets, have powerful capabilities that enable interpreting the large quantity and variety of patient data present across RPM and EHR. Leveraging GenAI technologies like Natural Language Processing to summarize, synthesize and navigate patient data and assist clinicians in documentation offers a promising pathway to address the critical challenge of clinician information overload [23].

*A. Core LLM capabilities relevant to integrated healthcare data analysis.*

LLMs bring a suite of capabilities uniquely suited to working with combined RPM and EHR data, enabled by their underlying architecture:

- **Natural Language Understanding (NLU) and Generation (NLG)**: The attention mechanisms in Transformer models allow LLMs to weigh the importance of different words (tokens) in context and their large parameter counts encode vast linguistic knowledge [9]. LLMs are able to generate fluent, contextually appropriate text in natural language, forming the basis for natural conversational interfaces [34] and automated report generation from analyzed data. These capabilities enable a nuanced understanding of complex clinical language and clinician queries.
- **Handling Diverse Inputs**: Alongside the native processing of text tokens, LLMs can be adapted for multimodal inputs [33][35]. Inherently multimodal architectures like Gemini, Med-PaLM use specialized encoders like Vision Transformers for images, and potentially others for time-series and fusion mechanisms like cross-attention to allow the model to jointly reason across different data types. Structured EHR data and numerical features derived from time-series RPM data can be tokenized or embedded into formats the LLM can process [10].
- **Information Synthesis and Summarization**: LLMs can process long sequences of input tokens and perform abstractive summarization, generating novel sentences that capture the essence of the input, rather than just extracting key phrases. The input tokens can include both EHR text [30] and appropriately encoded RPM data. This allows for more concise and readable summaries of potentially lengthy and complex integrated patient histories.
- **Reasoning and Question Answering**: LLMs develop rudimentary reasoning capabilities from the patterns in their massive training data [15]. Techniques like Chain-of-Thought (CoT) prompting [51] encourage step-by-step reasoning, improving performance on complex queries. This enables LLMs to formulate answers by grounding its reasoning in relevant facts [4] as shown by Liu et al (2023) [47].

*B. Prerequisites for Effective GenAI Applications in Healthcare*

Applying the capabilities of GenAI models to patient data safely, in order to reduce data overload and improve clinical efficiency, requires addressing a few key prerequisites.

*1) Robust Data Foundation*

- **Integration and standardization of data**: Healthcare data is often trapped in separate systems such as EHRs at different hospitals, labs, pharmacies, wearables that don't easily communicate or share information [14]. While GenAI doesn't solve the problem of interoperability, its potential drives the need for better integration between disparate data systems. A lack of standardization in EHR systems makes it hard to aggregate data from different sources for analysis. Common issues include inconsistent codes, terminologies and formats for the same diagnoses, medications, lab test results [2]. RPM data may not be in a standardized format across devices and raw data like sensor readings isn't inherently useful without the associated clinical context. Sophisticated analysis, often involving AI/ML, must be conducted to turn it into meaningful clinical insights by understanding context. To solve these issues, EHR data can be made accessible to the GenAI system's integration layer through standardized APIs such as Health Level 7 - Fast Healthcare Interoperability Resources (HL7-FHIR) as described by Tiase et al (2020) [43]. EHR data must be mapped across differing standards such as ICD (International Classification of Diseases), CPT (Current Procedural Terminology), and LOINC (Logical Observation Identifiers Names and Codes)[30]. Similarly, standardized outputs from RPM platforms are needed like IEEE 11073 - Personal Health Device (PHD) Standards [2]. Kawu et al (2023) [42] and Gene et al (2018) [54] describe a data integration framework to integrate PGHD from Apple HealthKit with the Epic EHR system.
- **Quality & Cleaning**: Information in EHR can be incomplete (gaps in patient history), inaccurate (typing errors, incorrect entries), inconsistent (mismatched terminology), or outdated [28][22]. RPM data from consumer devices may lack the rigorous validation of clinical-grade equipment, thereby impacting reliability [1]. Data continuity may be affected by device connectivity and battery life limitations. Patient-Reported Outcomes are highly dependent on patient engagement, tech literacy and manual data entry, often lacking context, and a reliance on snapshots of subjective data [21]. While primarily a data engineering task, GenAI can potentially assist in data cleaning by identifying inconsistencies and anomalies. GenAI can also be used for imputing missing values. Data preprocessing is required for consistent input quality of the data used in the LLM.
- **Privacy & Security:** Patient data is highly sensitive and legally protected (e.g., HIPAA, GDPR) [29]. Adopting cybersecurity best practices, managing patient consent, and anonymizing data are major concerns. Architectural solutions like secure HIPAA-compliant cloud environments, strong encryption and strict access controls must be adopted along with privacy-enhancing technologies like federated learning which involves training models locally without centralizing raw data or differential privacy during fine-tuning [36][29].

2) *Model Validation & Safety*
- **Accuracy**: Models need testing against established medical benchmarks such as MedQA for question answering [4]. As shown by Tam et al (2024) [46] evaluation metrics such as ROUGE (Recall-Oriented Understudy for Gisting Evaluation) [37], BERTScore (Score for Bidirectional Encoder Representations from Transformers) [37], AUROC (Area Under the Receiver Operating Characteristic) [53] must be considered along with benchmarks such as HELM (Holistic Evaluation of Language Models) [46]. Clinical validation studies to compare system outputs to expert judgment and patient outcomes is essential [31]. Grounding techniques must be adopted in the models to improve factuality. For clinical factuality, Retrieval Augmented Generation (RAG) [39] is crucial, allowing the LLM to access and cite specific, retrieved patient data from the integrated RPM/EHR store or external medical knowledge bases.
- **Domain Specificity**: LLMs must be fine-tuned using curated RPM and EHR datasets containing scenarios and clinical guidelines that adapt to the specific language and reasoning patterns of the domain [32]. RAG can also be used to provide access to up-to-date information without constant retraining.
- **Bias Mitigation**: Bias mitigation techniques must be used at every stage of training the models. Examples include auditing tools, balancing datasets during data preparation, using algorithmic fairness constraints during model training, and using fairness metrics across subgroups during output evaluation. [12].
- **Explainability**: Using Explainable AI (XAI) methods for LLMs such as interpreting attention weights, using simpler proxy models, generating natural language explanations via techniques like CoT along with source attribution via RAG contributes towards transparency [18].

3) *Adoption & Clinical Trust*
- **Workflow Integration**: GenAI-powered conversational interfaces should be designed to fit better into existing clinician workflows than static reports [16][8]. Integration with EHRs via APIs (like FHIR SMART) [43] help the conversational AI agent to be embedded within existing clinician tools.
- **Trust**: Demonstrated reliability, validation, transparency regarding limitations, insights from explainability methods, and user-centered design of the conversational interface are key factors in fostering clinicians' trust in AI assistance systems [7]. **Ethical and regulatory considerations:** The architecture must incorporate robust logging and audit trails to track AI inputs, outputs, and user interactions. This helps build accountability into the system. Compliance must be maintained with regulations around patient data and FDA guidance on AI-based SaMD (Software as a Medical Device), particularly regarding risk classification and required validation levels [13]. Consent mechanisms must be built into the user interface and data governance framework.

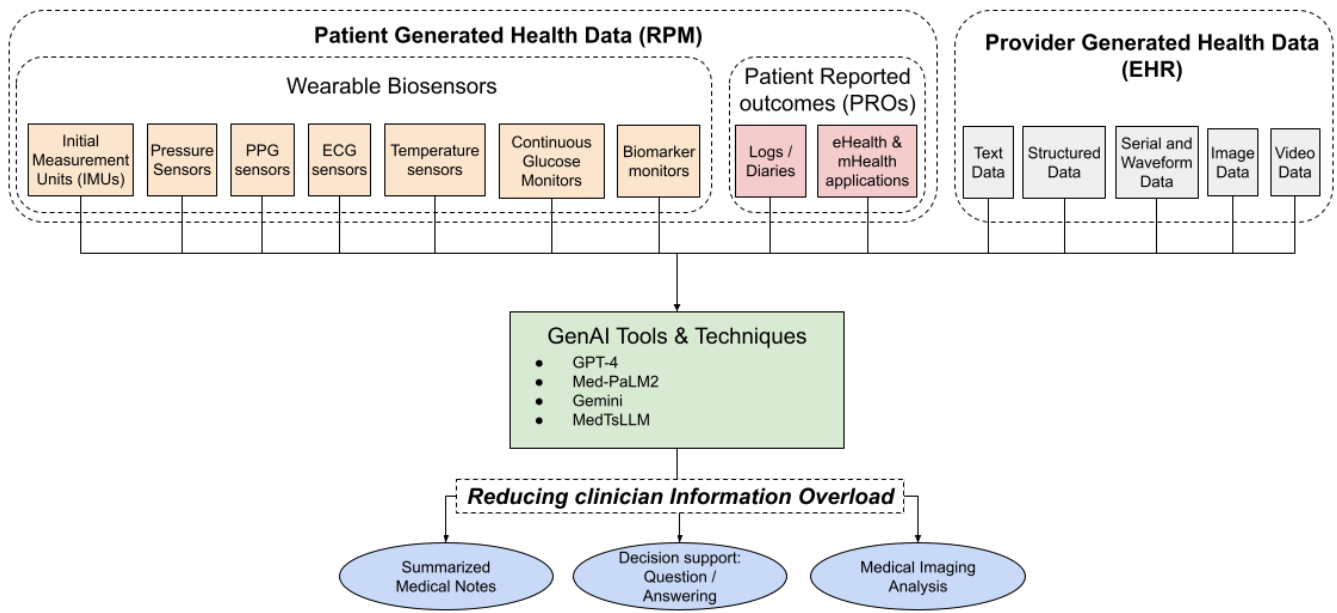

Figure 1: Leveraging GenAI techniques for managing complex patient health data to reduce clinical data overload

*C. Key Application Areas for GenAI with Integrated RPM/EHR Data*

GenAI models enable specific applications for reducing data overload for clinicians, improving workflows, assisting decision making and reducing documentation overhead:

*1) Enhancing Clinical Workflow Efficiency: Reducing Data Overload and Documentation burden*

As discussed in Section II, information overload is a major challenge for clinicians, in both, consuming and interpreting large amounts of patient data, and in generating detailed clinician notes [56]. To reduce the cognitive load, GenAI applications can be deployed for interacting with patient data:

*a) Text Summarization using LLMs*

One of the powerful applications of LLMs can be the summarization of EHR records to ease the information overload on clinicians. Lee et al (2024) [3] explore the potential of LLMs in summarizing patient charts and radiology reports, and its implications in reducing information overload. The review conducted by Li, Zhou et al (2024) [30] describes the use of evaluation metrics such as ROUGE, BERTScore, AUROC against data sources like MIMIC (Medical Information Mart for Intensive Care), CCKS (China Conference on Knowledge Graph) and i2b2/n2c2 (National NLP Clinical Challenges) to quantify the performance of LLMs. Zhang et al (2023) [25] have reviewed ongoing efforts to use LLMs in a variety of applications that help in reducing cognitive load for clinicians through support for administrative tasks and patient engagement with Natural Language Generation

*b) Note Generation using LLMs.*

Documentation overhead is another source of cognitive load that leads to workflow inefficiency. By virtue of their natural language generation abilities, LLMs are a great fit for clinical note generation. Discharge notes containing diagnosis, treatments and follow-up care can be generated through LLMs as shown through a qualitative and quantitative analysis conducted by Jung, HyoJe, et al. (2024) [40] for LLMs such as TinyLlama, LLama2, Mistral, BioMistral, Meditron, SOLAR.

Barak-Corren et al (2024) [50] have also shown quantifiable results in the use of ChatGPT that show improvements in clinician time spent on note-taking and charting, and a reduction in self-reported clinician effort in these tasks.

*2) Clinical Decision Support*

Conversational data querying using LLMs can be a valuable tool to support clinicians in decision-making. [24]. Conversational agents can reduce or eliminate the need for clinicians to use separate portals and manually review data, thereby improving the efficiency of established clinician workflows.

The potential of LLM in knowledge intensive clinical applications is supported by their ability to perform exceptionally well on complex medical tests. Bicknell et al (2024) [6] showcase GPT-4o achieving over 90% accuracy on 750 USMLE-style questions, against a medical student average of 59.3% with 95% Confidence Interval.

As shown by evaluations conducted by Singhal, Karan, et al. (2023) [52], and Nori, King et al (2023) [51], Large language models (LLMs) such as Flan-PaLM, Med-PalM 2, GPT-4 have exhibited a remarkable ability to interpret a wide array of domains and generate responses with natural language as it relates to the field of medicine and healthcare. In various performance benchmarks run on proficiency examinations containing multiple-choice and long-form medical question answering using popular datasets such as USMLE, MedQA, PubMedQA, MedMCQA, MMLU, LLMs have shown promising performance in physician-level medical question answering.

While this suggests that LLMs show strong medical knowledge required for clinical reasoning, further assessments beyond standardized tests are required to ensure clinical readiness. Applying raw knowledge contained in LLMs effectively in clinical workflows [8] requires frameworks for grounding LLM responses and mitigating hallucination risks.

To enhance conversational data querying in the healthcare context, grounding techniques such as Retrieval-Augmented Generated (RAG) can be used. Gao et al (2024) [39] have shown that GPT-4-turbo is able to match clinical recommendations made by physicians for antibiotic requests but not in other tasks.

"EMERGE" - a RAG driven framework proposed by Zhu, Ren et al (2024) [53] and REALM - a RAG framework proposed by Zhu et al (2024) [32] showcase methods for extracting insights from multimodal data including both time-series data and clinical notes in EHR systems. The process typically involves: (1) Clinician query (NLU by LLM). (2) LLM identifies key entities/concepts. (3) Retrieval-Augmented Generation (RAG) component converts these to embeddings/queries for a database containing indexed, preprocessed RPM/EHR data and extracted features. (4) Relevant data chunks retrieved. (5) LLM receives original query + retrieved context. (6) LLM generates a factually grounded answer (NLG). Elgedawy et al. (2024) [26] shows the use of RAG to facilitate a conversational interface, enabling a question-answer format of interaction for clinical notes stored in EHRs.

LLMs can also be used in RPM applications to work with time-series multi-modal PGHD gathered through sensor devices. Chan et al (2024) [10] have shown promising results with their MedTsLLM architecture, which focuses on semantic segmentation, boundary detection, and anomaly detection in respiratory and ECG data. Liu et al (2023) [47] demonstrate that LLMs have potential to be used with time-series physiological and behavioral data from wearable and clinical-grade sensing devices for varied tasks such as activity recognition, computing calories burned and atrial fibrillation classification with additional grounding and tuning of the LLMs.

Belyaeva, Cosentino, et al (2023) [35] define the HeLM framework: Health Large Language Model for Multimodal Understanding for training LLMs on individual-specific data, tokenizing and combining EHR and RPM data.

Wan et al (2024) showcases the application of LLMs for report generation based on multi-modal ECG data.[48] Wang, et al. (2024) [49] explore GPT-4's capabilities in generating structured diagnosis and treatment reports based on Thyroid cancer ultrasound description reports as a supplementary decision support tool.

## IV. Discussion

The preceding sections have highlighted the confluence of two major trends in healthcare: the exponential growth of patient data from RPM through ubiquitous wearable biosensors alongside traditional EHRs (Section II), and the emergence of powerful GenAI models, particularly Large Language Models, capable of sophisticated data synthesis and natural language understanding (Section III). We explored the significant challenge this data deluge poses for clinicians, particularly the problem of information overload and lack of efficient access to the data which leads to a lack of actionable insights. We examined the potential for GenAI systems to serve as an intuitive interface for navigating and interpreting this complex, integrated data landscape in Section III.

In this section we discuss the broader implications, challenges, and future directions for leveraging GenAI to bridge the gap between raw patient data and meaningful clinical application.

### A. Implications of a GenAI Approach

Adopting GenAI to interact with PGHD and EHR data holds considerable promise for transforming clinical practice. The primary implication is the potential to significantly reduce clinician cognitive load and improve efficiency [23]. The approach of allowing natural language querying and providing automated, context-aware summaries along with synthesizing trends across both real-time physiology/behavior (wearables) and clinical history (EHR), can drastically cut down time spent on manual data retrieval [1][20]. Moreover, LLM's have the capability to identify subtle patterns within this combined longitudinal data which could be missed by humans. This can help flag early signs of disease or suggest complex treatment responses that might otherwise be missed [4][20]. This moves towards enabling more proactive and personalized care, tailoring insights to the individual's unique, continuously monitored trajectory within their broader clinical context [35][55]. Ultimately, making complex data readily accessible through intuitive dialogue could democratize data interpretation and support more informed decision-making at the point of care.

### B. Challenges in Implementation

Despite the potential, realizing the benefits of this approach faces significant real-world hurdles, echoing challenges discussed in Sections II and III:

- **Foundational Data Integration**: Bridging technical and semantic gaps between diverse wearables and siloed EHRs to create a standardized, reliable data stream is a major prerequisite challenge requiring effort beyond the GenAI model itself [14][18][2].
- **Trustworthiness of GenAI in Clinical Contexts**: Ensuring the clinical validity, safety, and reliability of LLM-generated insights is paramount. Current problems include mitigating model hallucinations, ensuring factual grounding (where techniques like RAG, discussed in Section III, become critical), managing inherent biases learned from training data [12], and providing sufficient transparency or explainability for clinical acceptance [18]. Rigorous validation methodologies suitable for generative and conversational systems in high-stakes clinical environments are still maturing [31][22].
- **Privacy, Security, and Consent**: Handling highly sensitive, integrated health data via cloud-connected AI systems necessitates a strong focus on security measures and strict compliance with privacy regulations like HIPAA [29]. Current challenges involve managing granular patient consent for diverse data uses, ensuring secure data transmission and processing pipelines, and mitigating re-identification risks, especially as models become more powerful [31].
- **Workflow Compatibility and Adoption**: Care must be taken to ensure minimal disruption to clinical workflows. To build clinicians' trust in the system, conversational interfaces must be efficient, and integrate seamlessly with existing tools (like the EHR). Clear communication about the AI's limitations and an emphasis on feedback-driven human-centered design is crucial for adoption [7].

*C. Future Directions*

The successful adoption of conversational GenAI in clinical applications requires ongoing innovation:
- **Longitudinal Adaptation and Continual Learning:** Enabling models to continuously adapt to individual patient changes over time and be able to incorporate new medical knowledge without extensive retraining or compromising safety is essential for long-term sustainable adoption in clinical applications [36].
- **Efficiency and Scalability**: Developing LLMs with lower requirements for computing resources along with higher efficiency in inference techniques is crucial for maintaining cost-effectiveness at scale within healthcare systems.
- **Human-AI Collaboration Dynamics**: Further investigation and research is needed to optimize conversational interaction by understanding and adapting the user experience to be centered around clinician workflows. Verifying the quality of AI-generated information is also an important open problem [7].
- **Evolving Regulatory and Ethical Frameworks**: Creating validation approaches and ethical guidelines to keep pace with rapid GenAI advancements and ensure responsible innovation remains an ongoing societal and regulatory challenge in the healthcare domain [13].

## V. CONCLUSION

The concept of using LLMs for integrated data analysis offers distinct potential advantages over alternative solutions aimed at mitigating data overload such as dashboards and other GUI tools. In this paper, we present a comprehensive overview of the applications of GenAI techniques to improve clinician engagement with integrated patient data, focusing on the combination of real-time RPM [5] streams and traditional EHRs. We discuss various categories and modalities of patient data. Further, we discuss the capabilities, prerequisites and applications of a GenAI-approach in providing clinical insights from integrated RPM and EHR data. Lastly we discuss the implications, challenges in implementation and future direction for adoption of GenAI in clinical applications. With this review, we aim to guide future innovation in developing clinically relevant decision support tools that effectively leverage this complex data landscape.